\pdfoutput=1

\documentclass[11pt]{article}

\usepackage[]{acl}

\usepackage{times}
\usepackage{latexsym}
\usepackage{float}

\usepackage[T1]{fontenc}

\usepackage[utf8]{inputenc}

\usepackage{microtype}
\usepackage{hyperref}
\usepackage{url}
\usepackage{xspace}
\usepackage{amsmath}
\usepackage{amsthm}
\usepackage{amssymb}
\usepackage{booktabs}
\usepackage{multirow}
\usepackage{multicol}
\usepackage{ifthen}
\usepackage{graphicx}
\usepackage{multirow}
\usepackage{subcaption}
\usepackage[T1]{fontenc}
\usepackage{pifont}%
\usepackage[scaled=.85]{beramono}
\usepackage{colortbl}
\usepackage{subcaption}
\usepackage{enumitem}
\usepackage[capitalise,noabbrev]{cleveref}
\usepackage[most]{tcolorbox}
\usepackage{xcolor}

\newtcolorbox{mytextbox}[1][]{%
  enhanced,
  colback=white,
  height=10cm,
  attach title to upper,
  #1
}

\usepackage{microtype}

\usepackage{amsmath,amsfonts,bm}

\def\eqref#1{equation~\ref{#1}}

\def\1{\bm{1}}

\DeclareMathAlphabet{\mathsfit}{\encodingdefault}{\sfdefault}{m}{sl}
\SetMathAlphabet{\mathsfit}{bold}{\encodingdefault}{\sfdefault}{bx}{n}

\newcommand\todo[1]{\textcolor{red}{[TODO: #1]}}

\newcommand\hh[1]{\textcolor{blue}{[HH: #1]}}
\newcommand\nj[1]{\textcolor{brown}{[NJ: #1]}}
\newcommand\xp[1]{\textcolor{green}{[XP: #1]}}

\renewcommand\hh[1]{}
\renewcommand\todo[1]{}
\renewcommand\nj[1]{}
\renewcommand\xp[1]{}

\newcommand\ie{i.e.\xspace}
\newcommand\eg{e.g.\xspace}

\newcolumntype{L}[1]{>{\raggedright\let\newline\\\arraybackslash\hspace{0pt}}m{#1}}
\newcolumntype{C}[1]{>{\centering\let\newline\\\arraybackslash\hspace{0pt}}m{#1}}
\newcolumntype{R}[1]{>{\raggedleft\let\newline\\\arraybackslash\hspace{0pt}}m{#1}}

\ifthenelse{\isundefined{\definition}}{\newtheorem{definition}{Definition}}{}
\ifthenelse{\isundefined{\assumption}}{\newtheorem{assumption}{Assumption}}{}
\ifthenelse{\isundefined{\hypothesis}}{\newtheorem{hypothesis}{Hypothesis}}{}
\ifthenelse{\isundefined{\proposition}}{\newtheorem{proposition}{Proposition}}{}
\ifthenelse{\isundefined{\theorem}}{\newtheorem{theorem}{Theorem}}{}
\ifthenelse{\isundefined{\lemma}}{\newtheorem{lemma}{Lemma}}{}
\ifthenelse{\isundefined{\corollary}}{\newtheorem{corollary}{Corollary}}{}
\ifthenelse{\isundefined{\alg}}{\newtheorem{alg}{Algorithm}}{}
\ifthenelse{\isundefined{\example}}{\newtheorem{example}{Example}}{}
\newcommand*\diff{\mathop{}\!\mathrm{d}}

\title{Are All Spurious Features in Natural Language Alike? An Analysis through a Causal Lens}

\author{
Nitish Joshi$^{1}\thanks{\:\:equal contribution}$ \ \ \ \ \ Xiang Pan$^{1*}$ \ \ \ \ \ He He$^{1,2}$  \\
 $^1$Department of Computer Science, New York University\\
 $^2$Center for Data Science, New York University \\
  {\texttt{\{nitish, xiangpan, hhe\}@nyu.edu}} \\
}

\begin{document}
\maketitle
\begin{abstract}
    The term `spurious correlations' has been used in NLP to informally denote any undesirable feature-label correlations.
However, a correlation can be undesirable because
(i) the feature is irrelevant to the label (\eg punctuation in a review),
or (ii) the feature's effect on the label depends on the context (\eg negation words in a review),
which is ubiquitous in language tasks.
In case (i), we want the model to be invariant to the feature, which is neither necessary nor sufficient for prediction.
But in case (ii), even an ideal model (\eg humans) must rely on the feature, since it is necessary (but not sufficient) for prediction.
Therefore, a more fine-grained treatment of spurious features is needed to specify the desired model behavior. 
We formalize this distinction using a causal model and probabilities of necessity and sufficiency,
which delineates the causal relations between a feature and a label.
We then show that this distinction helps explain results of existing debiasing methods on different spurious features,
 and demystifies surprising results such as the encoding of spurious features in model representations after debiasing.

\end{abstract}

\section{Introduction}
\label{sec:intro}

Advancements in pre-trained language models ~\cite{devlin-etal-2019-bert, radford2019language} and large datasets ~\cite{rajpurkar-etal-2016-squad, wang-etal-2018-glue} have enabled tremendous progress on natural language understanding (NLU). This progress has been accompanied by the concern of models relying on superficial features such as negation words and lexical overlap ~\cite{poliak-etal-2018-hypothesis, gururangan-etal-2018-annotation, mccoy-etal-2019-right}. 
Despite the progress in building models robust to spurious features ~\cite{clark-etal-2019-dont,he2019unlearn, Sagawa*2020Distributionally, veitch2021counterfactual,puli2022outofdistribution},
the term has been used to denote any feature that the model should not rely on, as judged by domain experts.

Our key observation is that a feature can be considered spurious for different reasons. 
Compare two such features studied in the literature (\cref{tab:eg}):
(a) director names (such as `Spielberg') in sentiment analysis ~\cite{wang-culotta-2020-identifying};
(b) negation words in natural language inference ~\cite{gururangan-etal-2018-annotation}.
We do not want the model to rely on the director name because removing or changing it does not affect the sentiment.
In contrast,
while models should not \emph{solely} rely on the negation word, they are still \emph{necessary} for prediction---it is impossible to determine the label without knowing its presence. 

\begin{table}[t]
\begin{small}
	\centering
	
	\begin{tabular}{l}
		\toprule
       
        \textbf{Irrelevant features} \\
        
        {\color{red}{Speilberg}}'s new film is brilliant. $\longrightarrow$ \color{blue}{Positive} \\
       
        \rule{0.9cm}{0.15mm}'s new film is brilliant. $\longrightarrow$ \color{blue}{Positive} \\
        \midrule
         
         \textbf{Necessary features} \\
       
        \textit{The differential compounds to a hefty sum over time.}\\
        
        The differential will {\color{red}{not}} grow $\longrightarrow$ \color{blue}{Contradiction}   \\
        
        The differential will  \rule{0.5cm}{0.15mm} grow $\longrightarrow$ \color{orange}{?} \\
        
        \bottomrule
	\end{tabular}%
	\vspace{-4pt}
	\caption{Difference between two spurious features:
	(a) the director name can be replaced without affecting the sentiment prediction;
	(b) the negation word is necessary as it is not possible to determine the label without it.
	}
	\label{tab:eg}%
\end{small}
\vspace{-0.5cm}
\end{table}%

In this work, we argue that many spurious features studied in NLP are of the second type where the feature is necessary (although not sufficient) for prediction, which is more complex to deal with than completely irrelevant features in the first case. 
Current methods do not treat the two types of feature separately,
and we show that this can lead to misleading interpretation of the results.

To formalize the distinction illustrated in \cref{tab:eg}, we borrow notions from causality \citep{wang2021desiderata, Pearl1999-PEAPOC}, and use probability of necessity (PN) and probability of sufficiency (PS) to describe the relation between a feature and a label.
Intuitively, high PN means that changing the feature is likely to change the label
(\eg removing ``not'' will flip the label);
high PS means that adding the feature to an example would produce the label
(\eg adding ``the movie is brilliant'' to a neutral review is likely to make it positive).
Under this framework, we define two types of spurious features (Section~\ref{sec:formalism}): {irrelevant features} (\eg the director name)
that have low PN and low PS, and {necessary features} (\eg the negation word) that have high PN despite low PS.

Next, we describe the challenges in evaluating and improving robustness to necessary spurious features (Section~\ref{sec:learning}).
First, necessary features compose with other features in the context to influence the label.
Thus, evaluating whether the model relies solely on the necessary feature requires perturbing the context. 
This process often introduces new features and leads to inconsistent results depending on how the context is perturbed.

Second, we analyze the effectiveness of two classes of methods---data balancing and representation debiasing---on the two types of spurious features.
Data balancing breaks the correlation between the label and the spurious feature (\eg \citet{sagawa2020investigation});
 representation debiasing directly removes the spurious feature from the learned representation (\eg \citet{ravfogel-etal-2020-null}).
Although they are effective for irrelevant features,
we show that for necessary spurious features,
(i) data balancing does not lead to invariant performance with respect to the spurious feature (Section~\ref{ssec:feature_label_ind});
and (ii) removing the spurious feature from the representation significantly hurts performance (Section~\ref{ssec:rep_invariance}).

In sum, this work provides a formal characterization of spurious features in natural language.
We highlight that many common spurious features in NLU are necessary (despite being not sufficient) to predict the label, which introduces new challenges to both evaluation and learning.

\begin{figure*}[t!]
    \centering
    \begin{subfigure}[b]{0.3\textwidth}
        \includegraphics[scale=0.8]{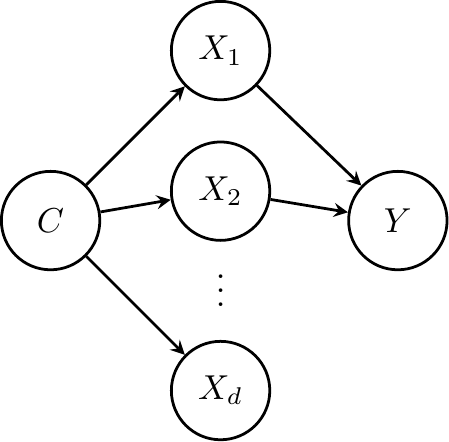}
        \caption{Data generating model.}
        \label{fig:data-generating}
    \end{subfigure}
    \hspace{1em}
    \begin{subfigure}[b]{0.3\textwidth}
        \includegraphics[scale=0.8]{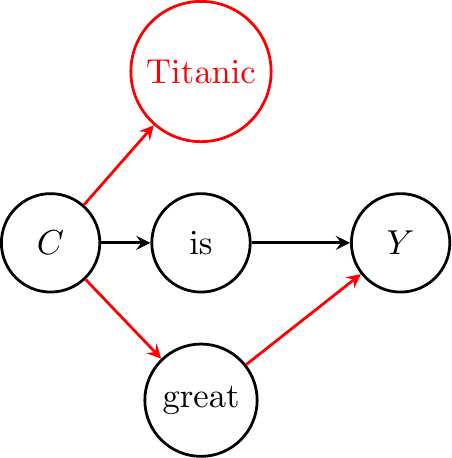}
        \caption{Type 1 dependence.}
        \label{fig:pure-spurious}
    \end{subfigure}
    \hspace{2em}
    \begin{subfigure}[b]{0.3\textwidth}
        \includegraphics[scale=0.8]{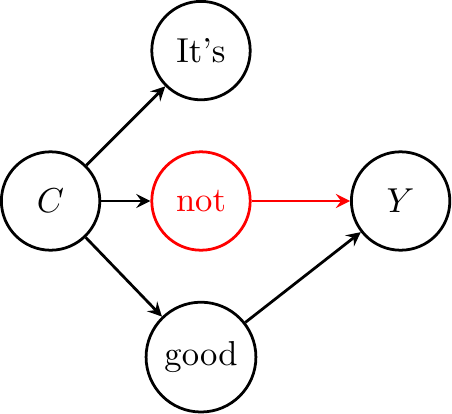}
        \caption{Type 2 dependence.}
        \label{fig:non-pure-spurious}
    \end{subfigure}
    \caption{
        Causal models for text classification. 
        (a) $C$ is the common cause of words in the input.
        Each word $X_i$ may be causally influence $Y$.
        (b) $Y$ (sentiment label) and $X_i$ (``Titanic'') are dependent because of the confounder $C$ (indicated by the red path).    
        (c) $Y$ (sentiment label) and $X_i$ (``not'') are dependent because of a causal relation.
    }
    \label{fig:model}
\end{figure*}

\section{Categorization of Spurious Features}
\label{sec:formalism}

\subsection{Causal Models}
\label{sec:causal_models}
We describe a structural causal model for text classification 
to illustrate the relation between different spurious features and the label. 
Let $X=(X_1, X_2,..,X_n)$ denote a sequence of input words/features\footnote{For illustration purposes, we assume that each feature is a word in the input text. However, the same model and analysis apply to cases where $X_i$ denote a more complex feature (e.g. named entities or text length) extracted from the input.}
and $Y$ the output label.
We assume a data generating model shown in \cref{fig:data-generating}.
There is a common cause $C$ of the input (\eg a review writer, a PCFG or a semantic representation of the sentence),
conditioned on which the words are independent to each other.
Each word $X_i$ may causally affect the label $Y$. 

Under this model, the dependence between $Y$ and a feature $X_i$ can be induced by two processes.
The \textit{type 1 dependence} is induced by a confounder (in this case $C$) influencing both $Y$ and $X_i$ due to biases in data collection,
\eg search engines return positive reviews for famous movies;
we denote this non-causal association by the red path in \cref{fig:pure-spurious}.
The \textit{type 2 dependence} is induced by input words that causally affect $Y$ (the red path in \cref{fig:non-pure-spurious}),
\eg negating an adjective affects the sentiment.
Importantly, the two processes can and often do happen simultaneously.
For example, in NLI datasets, the association between negation words and the label is also induced by
crowdworkers' inclination of negating the premise to create a contradiction example.

A type 1 dependence (``Titanic''-sentiment) is clearly spurious because the feature and $Y$ are associated through $C$ while having no causal relationship.\footnote{
    The two types of dependencies are also discussed in \citet{veitch2021counterfactual}, where the type 1 dependence is called ``purely spurious''.
}
In contrast, a type 2 dependence (``not''-sentiment) is not spurious per se---even a human needs to rely on negation words to predict the label.

Now, how do we measure and differentiate the two types of feature-label dependence?
In the following, we describe fine-grained notions of the relationship between a feature and a label,
which will allow us to define the spuriousness of a feature.

\subsection{Sufficiency and Necessity of a Feature}

We borrow notions from causality to describe whether a feature is a necessary or sufficient cause of a label~\citep{Pearl1999-PEAPOC, wang2021desiderata}.
Consider the examples in \cref{tab:eg}:
intuitively, ``not'' is necessary for the contradiction label because in the absence of it (\eg removing or replacing it by other syntactically correct words)
the example would no longer be contradiction;
in contrast, ``the movie is brilliant'' is sufficient to produce the positive label because adding the sentence to a negative review is likely to increase its sentiment score.
Thus, the feature's effect on the label relies on counterfactual outcomes.

We use $Y(X_i=x_i)$ to denote the \emph{counterfactual label} of an example had we set $X_i$ to the specific value $x_i$.\footnote{The counterfactual label $Y(X_i=x_i)$ is also commonly written as $Y_{x_i}$ \cite{10.5555/1642718} but we follow the notation in ~\citet{wang2021desiderata}}

\begin{definition}[Probability of necessity]
    The probability of necessity (PN) of a feature $X_i=x_i$ for the label $Y=y$ conditioned on context $X_{-i}=x_{-i}$ is
    \begin{align*}
        &\text{PN}(X_i{=}x_i, Y{=}y \mid X_{-i}{=}x_{-i}) \triangleq \\ & p(Y(X_i \neq x_i) \neq y \mid X_i{=}x_i, X_{-i}{=}x_{-i}, Y{=}y) \;.
    \end{align*}
\end{definition}
Given an example $(x,y)$, $\text{PN}(x_i, y\mid x_{-i})$\footnote{For notational simplicity, we omit the random variables (denoted by capital letters) when clear from the context.}
is the probability
that the label $y$ would change had we set $X_i$ to a value different from $x_i$.
The distribution of the counterfactual label $Y(X_i \neq x_i)$
is defined to be $\int p(Y(X_i))p(X_i\mid X_i\neq x_i) \diff X_i$.
This corresponds to the label distribution when we replace the word $x_i$  with a random word that fits in the context (\eg ``Titanic'' to ``Ip Man''). 
In practice, we can simulate the intervention $X_i \neq x_i$ by text infilling using masked language models~\cite{devlin-etal-2019-bert}.

\begin{definition}[Probability of sufficiency]
    The probability of sufficiency (PS) of a feature $X_i=x_i$ for the label $Y=y$ conditioned on the context $X_{-i}=x_{-i}$ is
    \begin{align*}
        &\text{PS}(X_i{=}x_i, Y{=}y \mid X_{-i}{=}x_{-i}) \triangleq \\ & p(Y(X_i{=}x_i) = y \mid X_i{\neq}x_i, X_{-i}{=}x_{-i}, Y{\neq}y) \;.
    \end{align*}
\end{definition}

Similarly, $\text{PS}(x_i, y\mid x_{-i})$ is the probability that setting $X_i$ to $x_i$ would produce the label $y$
on an example where $x_i$ is absent.
For example, PS of ``not'' for the negative sentiment measures the probability that a positive review will become negative had we added ``not'' to the input.

We note that both PN and PS are \emph{context-dependent}---they measure the counterfactual outcome of {individual} data points.
For example, while ``not'' has high PN for contradiction in the example in \cref{tab:eg},
there are examples where it has low PN.\footnote{Consider the premise ``The woman was happy'' and the hypothesis ``The woman angrily remarked `This will not work!'''.}
Similarly, there can be examples where the word ``Titanic'' has high PN.\footnote{For example in sentiment analysis, consider `This movie was on a similar level as Titanic'.}
To consider the average effect of a feature, we marginalize over the context $X_{-i}$:
\begin{align*}
    \text{PN}(x_i, y){\triangleq}{\int}{\text{PN}}(x_i, y{\mid}X_{-i})p(X_{-i}{\mid} x_i, y) \diff X_{-i}
    ,
\end{align*}
and similarly for PS. 

\begin{definition}[Spuriousness of a feature]
    The spuriousness of a feature $X_i=x_i$ for a label $Y=y$ is $1-\text{PS}(x_i, y)$.
    We say a feature is spurious to the label if its spuriousness is positive.
\end{definition}

Our definition of the spuriousness of a feature follows directly from the definition of PS,
which measures the extent to which a feature is a sufficient cause of the label (marginalized over the context $X_{-i}$).
Following this definition, a feature is non-spurious only if it is sufficient in \emph{any} context.
Admittedly, this definition may be too strict for NLP tasks as arguably the effect of any feature can be modulated by context,
making all features spurious.
Therefore, practically we may consider a feature non-spurious if it has low spuriousness (\ie high PS).

\begin{figure}[ht]
    \centering
    \includegraphics[width=0.4\textwidth]{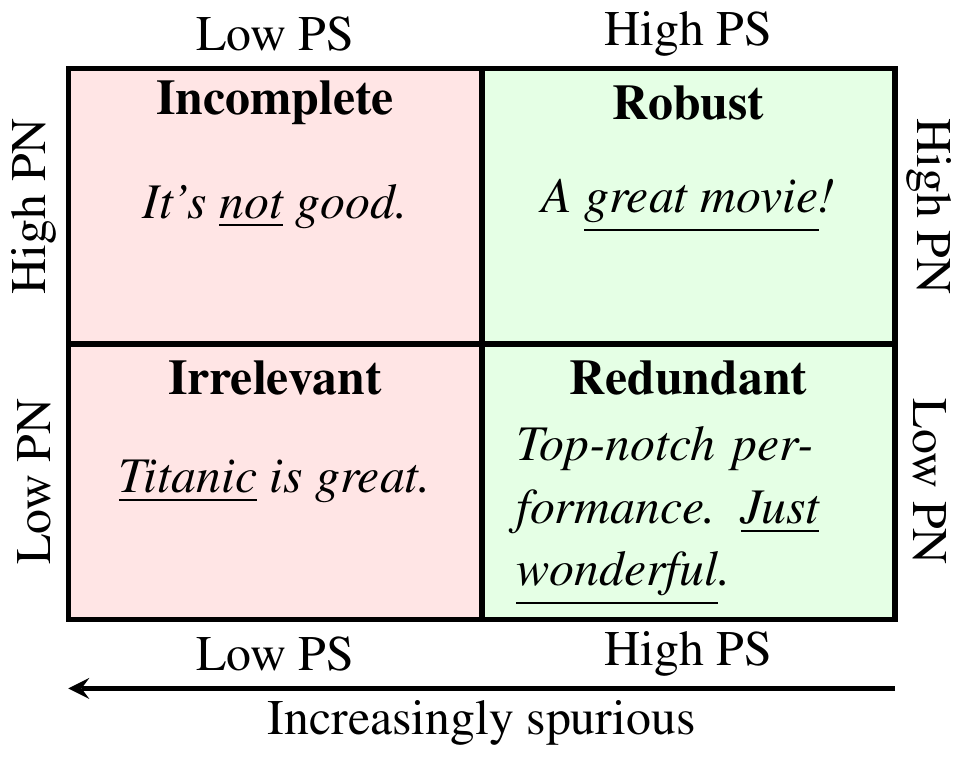}
    \caption{Categorization of \underline{features} based on their PN and PS. Spurious features have low PS. Among them, the high PN ones are part of
the features needed for prediction but they alone
are not sufficient;
and the low PN ones are irrelevant to prediction.}
    \label{fig:categories}
\end{figure}

\paragraph{Feature categorization.}
The above definitions provide a framework for categorizing features by their necessity and sufficiency to the label as shown in \cref{fig:categories}.

\paragraph{Estimating PN and PS.}
Calculating PN and PS of a feature requires knowing how the label would change when the feature is removed or added to an instance.
Sometimes we can reason about it with domain knowledge.
Consider the feature ``Titanic'' in \cref{fig:pure-spurious}, it has zero PN and PS since removing or adding it would not change the label.

In more complex cases, we might need to estimate the probabilities using an experiment.
For example, consider the lexical overlap between the premise and hypothesis in NLI. 
Given an entailment example with high word overlap,
changing the overlapped words is likely to cause label change (H1--3) unless it is replaced by a synonym (H4):\\[2mm]
\begin{small}
	\centering
	\begin{tabular}{ll}
		\textbf{P}: The doctor was paid by the actor.\\
		\textbf{H0}: The actor paid the doctor. & \textbf{L0}: Entailment\\
 		\midrule
		\textbf{H1}: The \textcolor{red}{teacher} paid the doctor. & \textbf{L1}: Neutral\\
		\textbf{H2}: The actor \textcolor{red}{liked} the doctor. & \textbf{L2}: Neutral\\
		\textbf{H3}: The actor paid the \textcolor{red}{guard}. & \textbf{L3}: Neutral\\
        \textbf{H4}: \textcolor{red}{An} actor paid the doctor. & \textbf{L4}: Entailment\\
	\end{tabular}%
\end{small}
\\[2mm]
Since a non-synonym is more likely to be sampled during intervention thus causing a label change,
we conclude that word overlap has high PN to entailment.
On the other hand, it is not a completely sufficient feature (i.e.\ spuriousness $> 0$) since there are plenty of examples with high lexical overlap but non-entailment labels \cite{mccoy-etal-2019-right}. 

We can partially automate this process by intervening examples using masked language models and then collecting labels for the perturbed examples.
We discuss this method in more detail and provide preliminary results in Appendix \ref{sec:estimating_pn}.
However, we note that while PN/PS can be estimated through careful intervention, as a conceptual framework, domain knowledge often suffices to judge whether a feature has high or low PN/PS.

\section{Experiment Setup}
Before diving into the implications of our categorization of spurious features,
we explain the common setup of experiments that we use to support our arguments.

\paragraph{Spurious features.}
Typical features considered in the literature (such as word overlap and negation words)
fall into the high PN and low PS category. 
Therefore, in the following discussion, we will focus on two types of spurious features:
low PN features that are irrelevant to prediction, and high PN features that are necessary but need additional context to decide the label. 

\paragraph{Datasets.} We use the following datasets that contain the spurious features.
(i) \textbf{Low PN spurious features}: we inject synthetic bias to MNLI examples
by associating a punctuation (`!!') with the neutral label.
Following \citet{dranker2021irm}, we set bias prevalence (i.e.\ examples where `!!' is present) to 25\% and set bias strength (i.e.\ percentage of examples with `!!' and the neutral label) to 90\%. The dataset is created by modifying MNLI examples through adding/deleting the feature at the end of the hypothesis.
(ii) \textbf{High PN spurious features}: we consider the negation bias ~\cite{poliak-etal-2018-hypothesis} and the lexical overlap bias in MNLI~\cite{williams-etal-2018-broad} for which we use the HANS challenge set~\cite{mccoy-etal-2019-right} during evaluation.

\paragraph{Models.} For all our experiments, unless otherwise stated, we use RoBERTa-large~\cite{Liu2019RoBERTaAR} from Huggingface \cite{Wolf2019HuggingFacesTS} as the backbone model.

\paragraph{Training methods.}
Our baseline algorithm
finetunes the pretrained model on the original dataset with cross-entropy loss.
We also experiment with debiasing methods including Subsampling \cite{sagawa2020investigation}, Product-of-Expert (POE) and Debiased Focal Loss (DFL) \cite{karimi-mahabadi-etal-2020-end} for comparison.
{Hyperparameters and training details can be found in Appendix~\ref{sec:exp_details}.}\footnote{Our code can be found at \url{https://github.com/joshinh/spurious-correlations-nlp}}

\section{Implications on Model Robustness}
\label{sec:learning}
Under the causal framework,
we say a model is \textit{non-robust}
if it fails on the interventional distribution.
For example, in \cref{fig:pure-spurious} the movie name has no causal effect on the label;
if intervening on it (\eg changing ``Titanic'') nevertheless incurs a prediction change,
we say the model is not robust.

\paragraph{Is relying on spurious features always bad?}
Prior work has suggested that if the model prediction relies on a single feature in any way, it is undesired \citep{gardner-etal-2021-competency}.
However, for a high PN feature, 
the label and the model output \textit{should} depend on it (\cref{fig:non-pure-spurious}).
Such dependency only becomes undesirable when other necessary features are ignored by the model
(\eg predicting negative sentiment whenever ``not'' is present).
This can be caused by two reasons:
first, the model may overly rely on a spurious feature $X_i$ due to confounding between $Y$ and $X_i$ in the training data (\eg ``not'' appears in all negative examples but not positive examples);
second, even without confounding, the model may fail to learn how $X_i$ interacts with other features to affect the label (\eg not understanding double negation).

\addtolength{\tabcolsep}{-2pt}
\begin{table}[t]
    \begin{small}
        \centering
        \begin{tabular}{lcrcr}
            \toprule
            Models        & \multicolumn{2}{c}{HANS} & \multicolumn{2}{c}{MNLI subsets}                  \\
                          & Ent/Non-ent   & $\Delta$                       & Ent/Non-ent & $\Delta$ \\
            \midrule
            BERT-base     & 99.2/12.9     & \cellcolor{red!86.3}{86.3}                        & 96.4/82.5 & \cellcolor{red!13.9}{13.9}    \\
            RoBERTa-large & 99.9/56.2  & \cellcolor{red!43.7}{43.7}                           & 97.1/93.6 & \cellcolor{red!3.5}{3.5}    \\
            \bottomrule
        \end{tabular}
        \caption{Results on two challenge sets for lexical overlap. Both indicate significantly different extent to which the models rely on the spurious correlation.
        }
        \label{tab:eval}
    \end{small}
\end{table}
\addtolength{\tabcolsep}{2pt}

\paragraph{How to evaluate models' robustness?}
A typical way to test models' robustness is to construct a ``challenge set''
that tests if perturbations of the input cause model predictions to change in an expected way.
The challenge here is that the expected behavior of a model depends on the type of the spurious feature. 
For low PN spurious features,
we can simply perturb them directly and check if the model prediction is invariant,
\eg replacing named entities with another entity of the same type \cite{balasubramanian-etal-2020-whats}.
Performance drop on this test set then implies that the model is non-robust.
However, intervention on the spurious feature only tells us if the feature is necessary, 
thus it cannot be used to evaluate robustness to high PN spurious features, where the model prediction is likely (and expected) to flip if we perturb the feature (\eg replacing ``not'' with ``also'' in \cref{fig:non-pure-spurious}).

For high PN spurious features like negation words, we instead want to test if they are sufficient for the model prediction.
An alternate method is to create two sets of examples with the same spurious feature but different labels.
For example, HANS \cite{mccoy-etal-2019-right} consists of entailment and non-entailment examples, both having complete lexical overlap;
this tests if high word overlap alone is sufficient to produce an entailment prediction. 
However, this process inevitably introduces a new variable.
Consider the causal graph in \cref{fig:non-pure-spurious}.
With the spurious feature (``not'') fixed, to change $Y$ we must change other features (\eg ``good'' $\rightarrow$ ``bad'') that affect the label by interacting with ``not''.
To make a correct prediction, the model
must learn the \emph{composite feature} formed by the spurious feature and the newly introduced features.
As a result, its performance depends not only on the spurious feature but also on
the features introduced during the perturbation. 

\paragraph{Inconsistent results on different challenge sets.}
To illustrate this problem, we evaluate models' robustness to lexical overlap on two challenge sets constructed differently: (a) HANS; (b) subsets of high lexical overlap examples in the MNLI dev set (where $>0.8$ fraction of words in the hypothesis are also in the premise).
Compared to (b),
HANS non-entailment examples require linguistic knowledge such as understanding passive voice (\eg ``The senators were helped by the managers'' does not imply ``the senators helped the managers'') or adverbs of probability (\eg ``Probably the artists saw the authors'' does not imply ``the artists saw the authors''), which are rare in MNLI.\todo{Maybe add quantitative results to show it is rare in MNLI.}

We fine-tune pre-trained models on MNLI and report their results in
Table~\ref{tab:eval}. 
While models perform poorly on high overlap non-entailment examples from HANS, their performance is much higher on such examples from MNLI (56.2\% vs 93.6\%),
leading to inconsistent conclusions.\footnote{Large variance in performance across different subcases of non-entailment examples as reported in ~\citet{mccoy-etal-2019-right} is another example of the unreliability.}
Thus, we should be careful when interpreting the magnitude of the problem on challenge sets, as the performance drop could also be attributed to unseen features introduced during dataset construction.

\section{Implications on Learning Methods}\label{sec:experiment}

In this section, we discuss two common classes of methods to train robust models and their effectiveness for spurious features with high/low PN. 

\subsection{Decorrelating the Spurious Feature and the Label}
\label{ssec:feature_label_ind}

A straightforward idea to remove undesirable correlation between the label $Y$ and a spurious feature $X_i$ due to confounding is to balance the training data 
such that $Y$ and $X_i$ are independent \cite{Japkowicz2000TheCI, austin2011introduction, li2019repair}. 
In practice, this amounts to subsampling the dataset to balance the classes conditioned on the spurious feature
(e.g. ``Titanic is good/bad'' are equally likely)
\cite{sagawa2020investigation},
or upweighting examples where the spurious feature is not predictive for the label \citep{karimi-mahabadi-etal-2020-end}.
While these methods have shown promise for spurious features with both high and low PN,
there is a key difference between the underlying mechanisms.

For a low PN spurious feature, the dependence between model prediction and the feature arises from a confounder that affects both $Y$ and $X_i$.
As shown in \cref{fig:pure-spurious}, assuming independence between the spurious feature and other features that affect the label
(\ie there is no path from $X_i$ to $Y$ through $C$),\footnote{While this is not true in general due to the complex grammar constraints in natural language, we use a simplified model for our analysis.}
$X_i$ and $Y$ are independent without confounding.
Thus, enforcing the independence through data balancing
matches the independence condition on the data generating distribution. 
As a result, the model prediction will be independent of $X_i$
and we expect its performance to be invariant across examples grouped by $X_i$ values (\eg similar accuracy on reviews about famous vs.\ non-famous movies).

On the other hand, for high PN spurious features, even without confounding,
$X_i$ is \emph{not} independent of $Y$ on the data generating distribution (Figure ~\ref{fig:non-pure-spurious}).
Then why do these methods work for high PN features?
Note that $X_i$ is not sufficient to decide $Y$ alone but forms a {composite feature} with other features that affect the label together (\eg a double negation construction). 
Therefore, within the same class, 
examples with different $X_i$ are likely to 
form different composite features.
In real data, certain combinations of $X_i$ and $Y$ (\eg positive examples with negation) often correlate with composite features that are difficult to learn (\eg double negation or comparison).
By balancing the $(X_i,Y)$ groups, we allow the model to learn the minority examples more effectively. 
However, the model performance is not necessarily invariant across groups because the model must rely on different (composite) features.

\begin{mytextbox}[height=1.4cm]Data balancing leads to invariance to low PN spurious features but not high PN ones.\end{mytextbox}

\paragraph{Experiments.}
We create balanced datasets for two spurious features in MNLI:
(a) punctuation, where examples are grouped by whether they end with `!!' as described in Section~\ref{sec:learning};
and (b) lexical overlap, where examples are grouped by lexical overlap (`high overlap' if more than $0.8$ fraction of the words in the hypothesis are also in the premise, and `low-overlap' if less than $0.2$).
For both groups, we subsample the training set such that the label distribution is uniform in each group.

To test models' invariance to groups, we train RoBERTa-large on one group and test on the other,
e.g. training only on high-overlap examples and evaluating on low-overlap examples --- a model that is invariant to the spurious feature should generalize equally well to both groups.

\begin{figure}[t]
    \centering
    \includegraphics[width=\linewidth
    ]{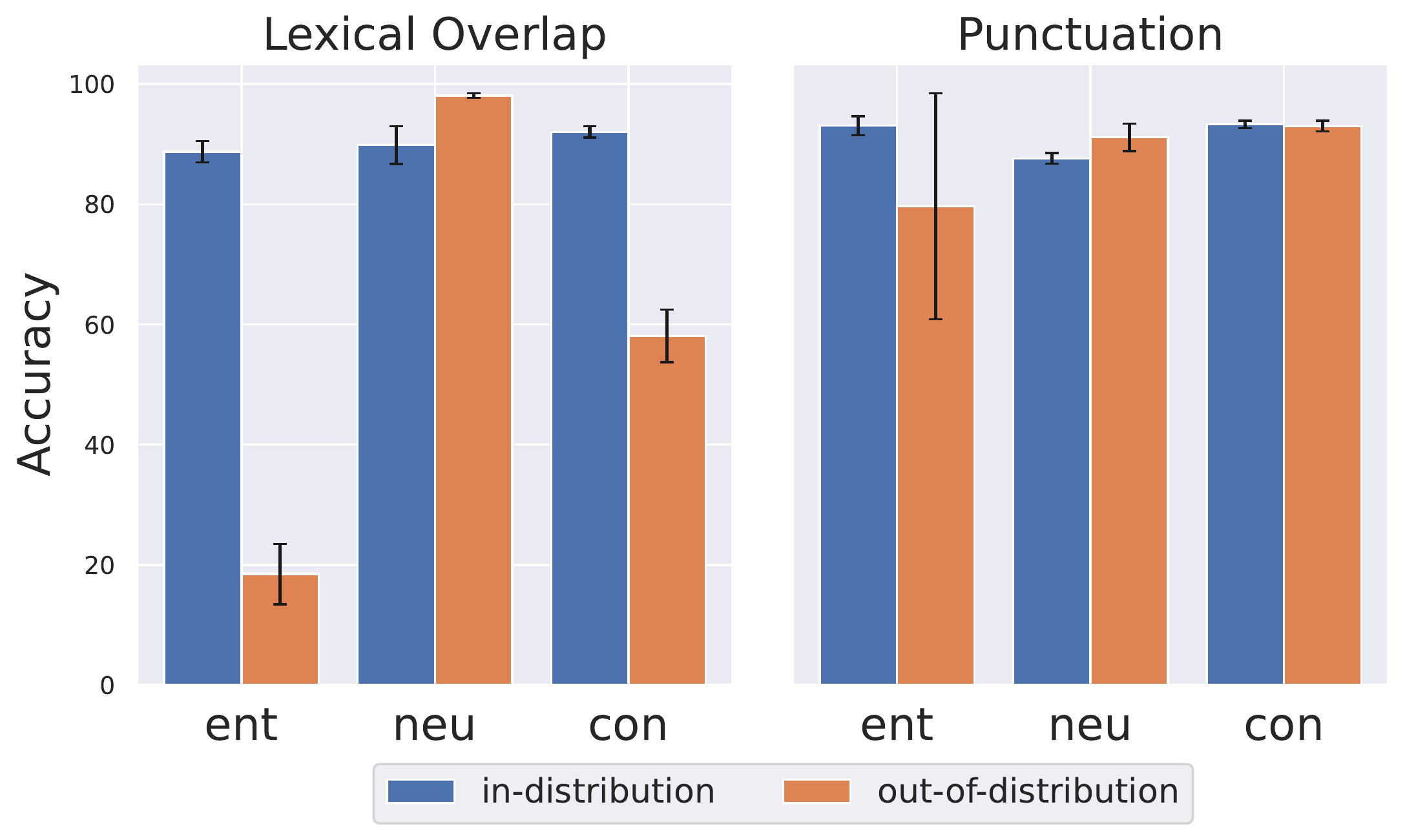}
    \caption{
    Model performance across groups.
    Left: train on high overlap examples.
    Right: train on examples with punctuation.
    We show both the \textcolor{blue}{in-distribution} performance where the model is tested on the same group as training,
    and the \textcolor{orange}{out-of-distribution} performance where the model is tested on the unseen group.
    Performance is invariant to groups if the feature has low PN (right) but has large variation if the feature has high PN (left).
    }
    \label{fig:group_generalization}
\end{figure}

\paragraph{Results.}
In Figure~\ref{fig:group_generalization}, we observe that for the punctuation feature (low PN), there is no large variance in performance across groups. 
But models have very different performances between the low and high overlap groups. Specifically, models trained on high overlap examples perform poorly on low overlap examples, in particular the entailment class, despite seeing no correlation between lexical overlap and label during training. This could happen because entailment examples within the high and low overlap groups require different features, such as lexical semantics in the high overlap group (``It stayed cold for the whole week'' implies `It stayed cold for the entire week''), and world knowledge in the low overlap group (``He lives in the northern part of Canada'' implies ``He stays in a cold place'') \cite{joshi-etal-2020-taxinli}.
The result highlights that for high PN spurious features, balancing the dataset might not be enough---we additionally need more examples (or larger models \citep{tu-etal-2020-empirical}) to learn the minority patterns.

\subsection{Removing Spurious Features from the Representation}
\label{ssec:rep_invariance}

A different class of methods focuses on removing the spurious feature from the learned representations, \eg iterative null-space projection \citep[][INLP]{ravfogel-etal-2020-null} and adversarial learning \cite{zhang2018mitigating}.
As argued in the previous section, high PN spurious features form composite features with other necessary features. Therefore, removing them also leads to the removal of the composite features,
which ends up hurting performance. 

\begin{mytextbox}[height=1.4cm]Removing high PN spurious features from the representation hurts performance.\end{mytextbox}

\paragraph{Experiments.}
We test our hypothesis by removing two spurious features (lexical overlap and punctuation) using INLP, a debiasing method that removes linearly encoded spurious features by iteratively projecting the learned representation.
We fine-tune RoBERTa-large on subsampled datasets where the label and the spurious feature are independent.
Over iterations of INLP, we measure the extractability of the spurious feature by its \emph{probing accuracy} and measure the model performance by its \emph{task accuracy},
where both are from linear classifiers trained on the debiased representations. Following ~\citet{mendelson-belinkov-2021-debiasing}, the linear classifiers are also trained and evaluated on balanced datasets.
For task accuracy, we report results on the minority group (e.g. high lexical overlap examples with non-entailment label) since we find that this group is most affected by debiasing.\footnote{Full results are in Appendix ~\ref{sec:inlp_extended}.}

\begin{figure}[t]
    \centering
    \includegraphics[scale=0.54]{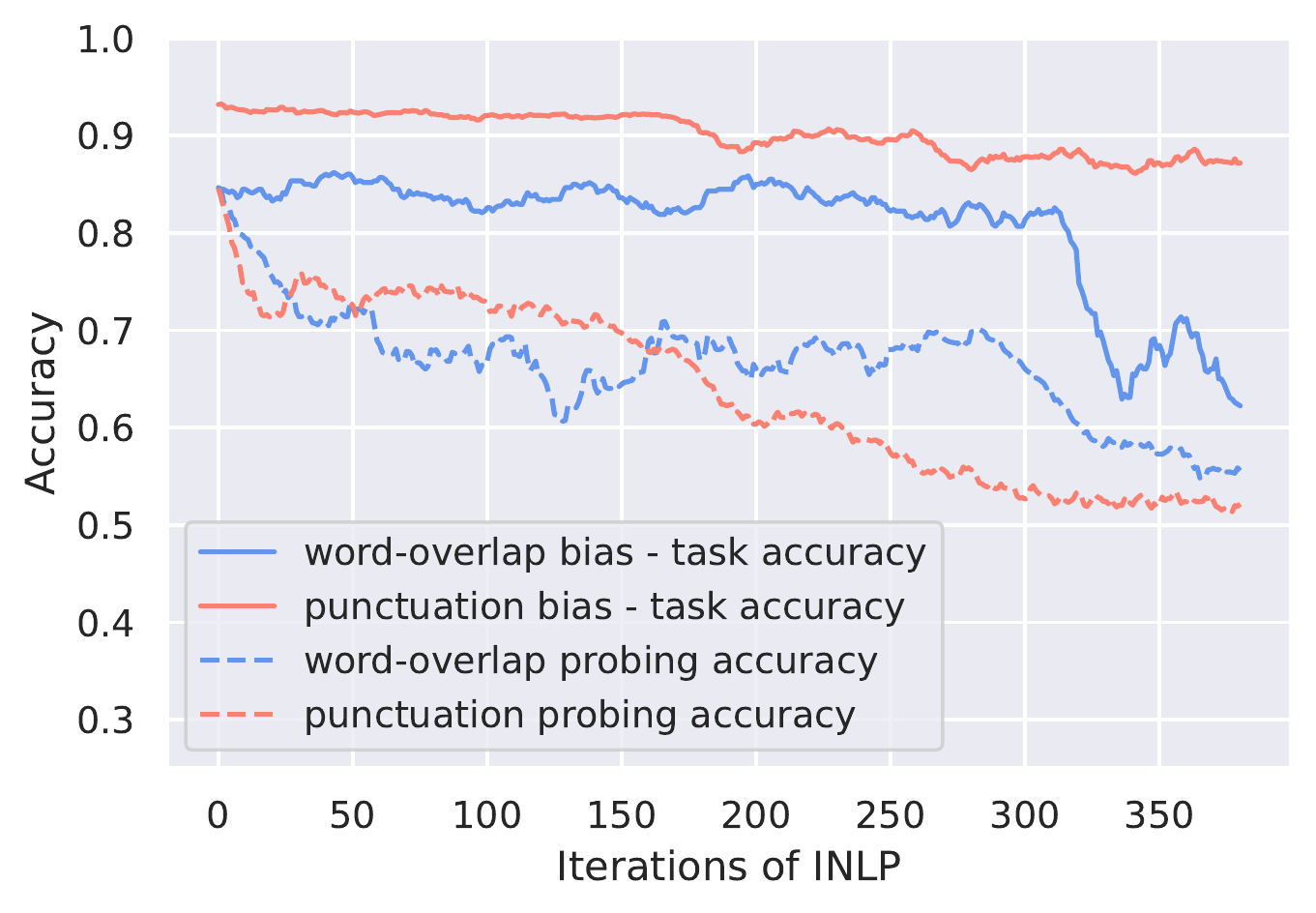}
    \caption{Extractability (probing accuracy) of the spurious feature (shown in dashed lines) and the task accuracy (shown in solid lines) as a function of iterations in INLP. For \textcolor{blue!80}{high PN features} (word-overlap), its removal (decreasing probing accuracy) is accompanied by large drop in the task accuracy.
    }
    \label{fig:inlp}
\end{figure}

\paragraph{Results.}
Figure~\ref{fig:inlp} shows the results for two spurious features. 
We observe that for high PN features (lexical overlap),
when the probing accuracy drops significantly around 300 iterations (i.e. the feature is largely removed from representation),
there is a significant drop in task accuracy.
In contrast, removing the low PN feature does not affect task accuracy significantly.

\subsection{What Features does the Model Learn with Data Balancing?}
We have seen that directly removing spurious features from the representation may hurt performance,
whereas data balancing generally helps.
Then what features do models learn from balanced data?
\citet{mendelson-belinkov-2021-debiasing} recently found that, quite counter-intuitively, it is easier to extract the spurious feature from the representation of models trained on balanced data. 
We argue that this occurs for high PN spurious features because they form composite features with other features, which a probe can rely on (\eg from ``not good'' we can still predict the existence of ``not''). 
In contrast, a low PN spurious feature that is not useful for prediction may become less extractable in the representation.

\begin{mytextbox}[height=1.4cm]Data balancing does not remove high PN spurious features from the representation.\end{mytextbox}

To understand the relation between a feature's correlation with the label (in the training set) and its prominence in the learned representation,
we first conduct experiments on a synthetic dataset where we can control the strength of feature-label correlations precisely.

\paragraph{Synthetic data results.} We create a binary sequence classification task similar to \citet{lovering2020predicting}, where each input is of length 10 from a vocabulary $V$ of integers ($|V| = 1k$).
We create spurious features with low and high PN as follows.
In the first task,
the label is 1 if the first two characters are identical;
the spurious feature is the presence of the symbol \texttt{2} in the sequence,
which has zero PN. 
In the second task, the label is 1 if the first two characters are identical XOR \texttt{2} is present; the spurious feature is again the presence of \texttt{2},
but in this case it has high PN (since removing \texttt{2} will flip the label). 

We generate a sequence of synthetic datasets with increasing bias strength by varying the correlation between the label and the spurious feature.
We then train LSTM models (embedding layer, a 1-layer LSTM and an MLP with 1 hidden layer with tanh activation) on each dataset and measure the extractability of the spurious feature from the model's representation.
Following \citet{mendelson-belinkov-2021-debiasing}, we train linear probes on balanced datasets to predict the feature from the last layer embeddings of each model.
We then measure extractability using two metrics: probing accuracy and compression $\mathcal{C}$ based on minimum description length~\cite{voita-titov-2020-information}.\footnote{For both metrics, higher value indicates higher extractability. See Appendix \ref{sec:exp_details} for more details about training.}

\begin{figure}
    \centering
    \includegraphics[scale=0.48]{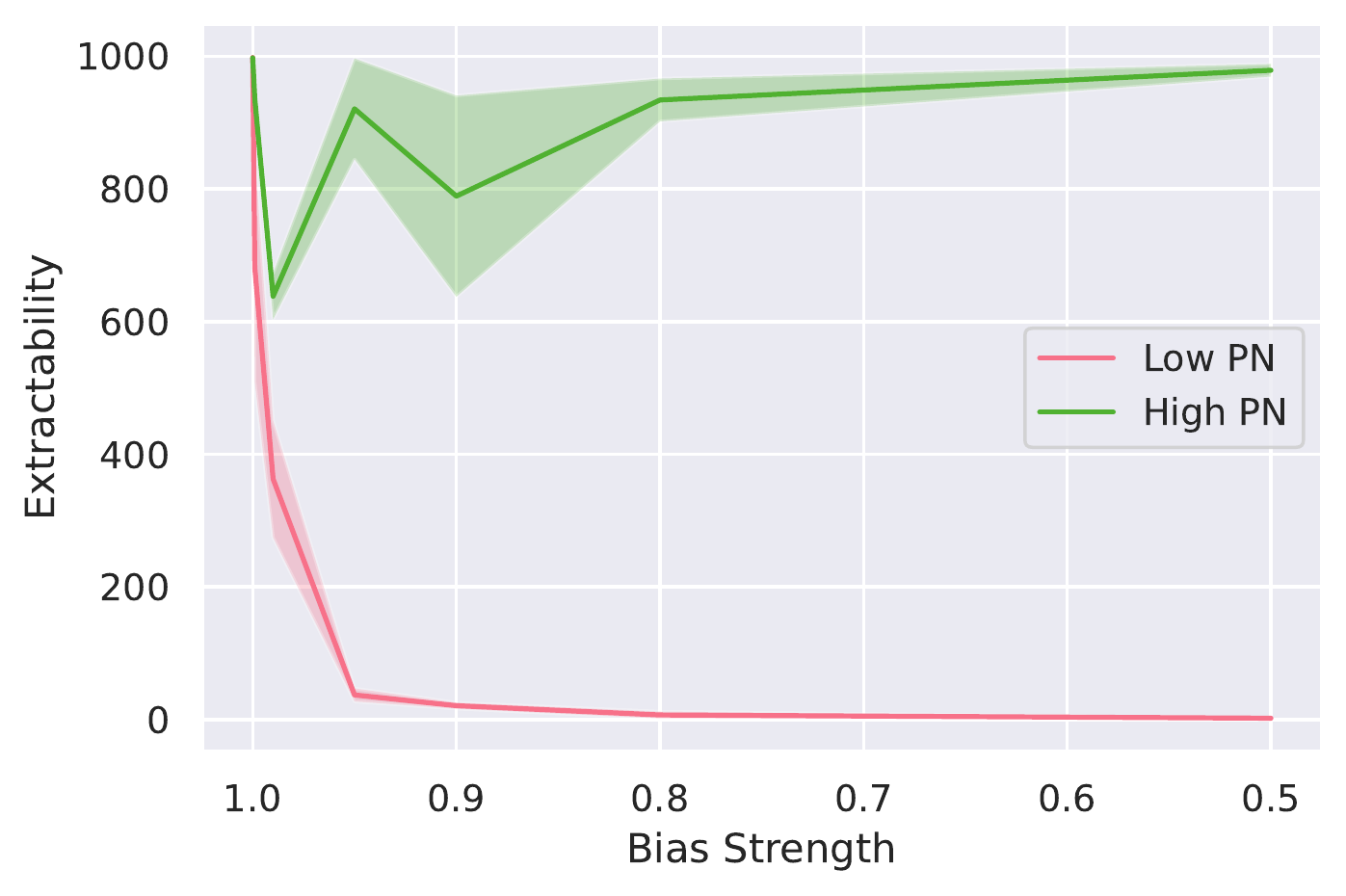}
    \caption{Extractability (compression) of the spurious feature as a function of bias strength on the synthetic data. The \textcolor{green!70!black}{high PN} feature is easily extractable regardless of its correlation with the label, whereas the  \textcolor{red!70}{low PN} feature becomes less extractable when the bias strength drops.}
    \label{fig:synthetic}
\end{figure}

\begin{table}[t]
\begin{small}
    \centering
    \begin{tabular}{lcccccc}
        \toprule
                   & \multicolumn{2}{c}{Lexical-overlap bias} & \multicolumn{2}{c}{Punctuation Bias}                               \\
                                            &         $\mathcal{C}$                     & Acc. & $\mathcal{C}$ & Acc. \\
        \midrule
        Baseline                                    & 3.5                              & 90.5 & 47.6          & 100.0  \\
        \midrule
        Subsampled                                    & 3.6                               & 91.5 & \cellcolor{red!74}{10.2}          & 97.7 \\
        POE                                    & \cellcolor{cyan!60}{4.2}                               & 91.3 & \cellcolor{red!9.4}{42.9}       &  99.9     \\
        DFL                                 & \cellcolor{cyan!30}{3.9}                               & 88.3 & \cellcolor{cyan!1.8}{48.5}              & 100.0     \\
        \bottomrule
    \end{tabular}
    \caption{Extractability of the spurious feature for various robust training methods. Blue denotes an increase whereas red denotes a decrease in extractability from the baseline. For high PN spurious features (lexical overlap), the feature is as easy if not easier to extract after debiasing, as compared to the baseline, in contrast to the low PN feature (punctuation).
    }
    \label{tab:probing}
\end{small}
\end{table}

Figure \ref{fig:synthetic} plots the extractability of the spurious feature (measured by compression $\mathcal{C}$) as a function of the bias strength.
We observe that the extractability of the high PN spurious feature remains high across varying bias strengths, including when the spurious feature and the label are independent (bias strength=0.5).
In contrast, for low PN spurious features, we observe that its extractability decreases as the bias strength decreases.
In other words, they become less prominent in the representation as their correlation with the label drops.

\paragraph{Real data results.}
Next, we study the effect of debiasing algorithms on the extractability of spurious features in real datasets.
We evaluate the following methods: Subsampling \cite{sagawa2020investigation}, Product-of-Expert (POE) and Debiased Focal Loss (DFL) \cite{karimi-mahabadi-etal-2020-end}, all of which explicitly or implicitly break the feature-label correlation during training. We also train using ERM on the original biased dataset as a baseline. All methods use RoBERTa-large as the backbone model.
We test on the low PN spurious feature (punctuation, `!!') and the high PN spurious feature (lexical overlap) in Table~\ref{tab:probing}.\footnote{The results for negation bias can be found in Appendix \ref{sec:encoding_extended}.} 

We observe that the high PN feature, lexical overlap, is still easily extractable after debiasing. 
In contrast, for the low PN feature, punctuation, although its probing accuracy is high, its compression is larger in the baseline models, i.e. the feature becomes harder to extract after debiasing, which is consistent with what we observe in the synthetic case.

In sum, we show that breaking the correlation between a feature and the label (\eg through data balancing) does not necessarily remove the feature from the learned representation.
The high PN features can still be detected from the composite features on which the label depends.

\section{Related Work}\label{sec:related}

While there is a large body of work on improving model robustness to spurious correlations,
the question of what spurious features are in natural language
is less studied.

\citet{veitch2021counterfactual} formalize spurious correlations from a causal perspective and argued that the right objective is counterfactual invariance (CI)---the prediction of a model should be invariant to perturbations of the spurious feature. They also make a distinction between purely spurious and non-purely spurious correlations, which are similar to the type 1 and type 2 dependencies we defined.
However, their main approach and results assumed purely spurious correlations.
Here, we argue that high PN features, or non-purely spurious correlations, are more common in NLP tasks,
and the label is not invariant to these features.

\citet{gardner-etal-2021-competency} consider all single features/words that correlate with the label as spurious. Under this definition, the learning algorithm should enforce a uniform distribution of the prediction conditioned on any feature, i.e. $Y | X_i=x_i$ should follow a uniform distribution (termed uninformative input features or UIF by \citet{Eisenstein2022UninformativeIF}). 
To connect PN/PS (counterfactual quantities) with the conditional probability (an observational quantity),
we must marginalize over the context.
If the feature has zero PN and PS (i.e. it has no effect on the label in any context),
$p(Y \mid X_i=x_i)$ is uniform for all $x_i$.
However, we cannot say the same for features with non-zero PN/PS. 

Recently, \citet{Eisenstein2022UninformativeIF} used a toy example to demonstrate the disconnect between UIF and CI, showing that neither objective implies the other. Along similar lines, ~\citet{Schwartz:2022} argued that UIF is hard to achieve in practice; further, enforcing a uniform label distribution for one feature may skew the label distribution for other features.
Our work complements the two by adding more clarity to the relation between a feature and the label in NLP tasks.
Additionally, we highlight that neither the CI nor the UIF principle holds for high PN spurious features,
which the label depends on in the true data generating distribution.

Finally, formal notions of \emph{necessity} and \emph{sufficiency} from causality have also been used in the context of explanations.
~\citet{Mothilal2021TowardsUF} and ~\citet{Galhotra2021ExplainingBA} use a causal framework and counterfactual examples to estimate necessity and sufficiency of explanations. ~\citet{wang2021desiderata} used the notions to formalize the desired properties of representations---they should be non-spurious (capturing sufficient features) and efficient (every feature should be necessary). We use notions of probability of causation to formalize two different types of spurious features present in natural language.

\section{Conclusion}
\label{sec:conclusion}

In this work, we showed that all spurious features in natural language are not alike---many spurious features in NLU are necessary but not sufficient to predict the label. We further showed how this distinction makes it challenging to evaluate model robustness and to learn robust models.
In particular, unlike low PN spurious features that are irrelevant to prediction,
high PN features interact with other features to influence the label.
Therefore, they do not have a clean relationship with the label that allows us to enforce independence or invariance during training.

Perhaps a pessimistic takeaway is that there is not much we can do about high PN spurious features.
The key problem is that the model fails to learn the rare or unseen compositions of the necessary spurious feature and other features (\eg different constructions that involve negation).
That said, we believe large language models suggest promising solutions because
1) they have good representations of various constructions in natural language;
2) they can bypass the problem of dataset bias in supervised learning through few-shot in-context learning;
3) they can take additional inductive bias for the task through natural language prompting (e.g. chain-of-thought).
We hope that our result will spur future work on training and evaluating spurious correlations that are more suited for spurious features arising in natural language.

\section*{Limitations}
\label{sec:limitations}

While our definition helps put spurious features into perspective, it has some limitations:
\begin{enumerate}
    \item Our definition relies on counterfactual quantities which are not observed. Thus, actually computing PN and PS is expensive and needs a human to, at the very least, go through the perturbed examples. 
    \item While the definitions and categorization help interpret experiment results, they do not directly tell us what training \& evaluation methods are suitable for the high PN spurious features in particular.
    One straightforward idea to enforce models to match the PN and PS of features in the data generating distribution. This would require collecting counterfactual examples with control for a specific feature (as opposed to generic counterfactuals as in ~\citet{kaushik2020learning}). We believe that more research is needed to understand how to train models robust to spurious correlations. Both our work and ~\citet{Schwartz:2022} argue that subsampling training data to ensure the independence between the spurious feature and the label might not work. Nevertheless, we believe that our definitions are important to put the results in perspective and make progress.
\end{enumerate}
\section*{Acknowledgements}

We thank Sameer Singh, Nicholas Lourie, Vishakh Padmakumar, Richard Pang, Chen Zhao, and Saranya Venkatraman for discussion and feedback on the work. We thank Yixin Wang for pointing out an error in our initial causal model. NJ is supported by an NSF Graduate Research Fellowship under grant number 1839302. This work is partly supported by Samsung Advanced Institute of Technology (Next Generation Deep Learning: From Pattern Recognition to AI) and a gift from AWS AI.

\bibliography{anthology,custom}
\bibliographystyle{acl_natbib}

\appendix

\clearpage

\section{Measuring PN}
\label{sec:estimating_pn}

To provide a more concrete method for measuring PN of any feature, we use the following method: We use masked language models (MLMs) \cite{devlin-etal-2019-bert} to intervene on the feature $X_i$ by masking and in-filling while ensuring that $x_i^\prime \neq x_i$ i.e. the replaced word is different from the original one. We can then annotate these examples (either using experts or through crowdsourcing) to check if the new label is the same. We use this method to compute PN over a small set of randomly sampled examples (20) which were annotated by the authors. We used RoBERTa-large for mask in-filling. Using this method, the estimated PN for negation features is 0.8, for lexical overlap it is 0.7 and for punctuation bias in NLI it is 0. This shows that, as expected, lexical overlap and negation features have much higher PN than punctuation. We note that while such a method is useful to estimate PN/PS, as a conceptual framework, domain knowledge often suffices to judge whether a feature has high or low PN/PS.

\section{Experimental Details}
\label{sec:exp_details}

In all the experiments, the model is trained for 3 epochs, with a maximum sequence length of 128 tokens. We use a learning rate of 1e-5 with the Adam Optimizer ~\cite{Kingma2015AdamAM} with a batch size of 32. All experiments were run on a single RTX8000 GPU with run-time of $<12$ hours for each experiment. We use the default train/dev split in MNLI dataset.

Probing Experiments (Section ~\ref{ssec:rep_invariance}): We use setting similar to ~\citet{mendelson-belinkov-2021-debiasing} where we train linear probes on subsampled datasets where the probing label is balanced. The probe is trained with a batch size of 64 for 50 epochs with a learning rate 1e-3 using Adam optimizer. 

\section{INLP: Extended Results}
\label{sec:inlp_extended}

\begin{figure}[t]
    \centering
    \includegraphics[scale=0.54]{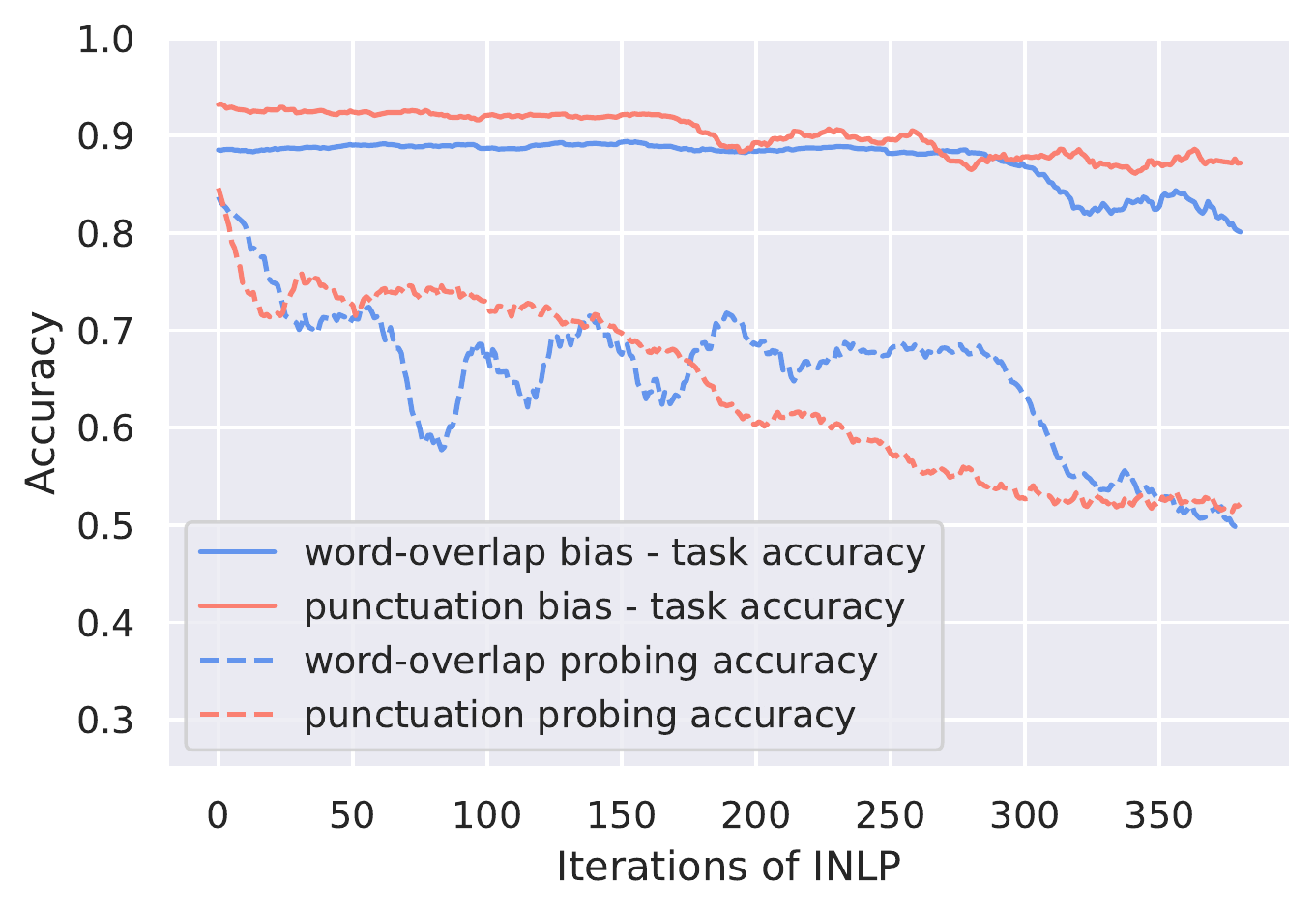}
    \caption{Extractability of the spurious feature (probing accuracy) and the main task accuracy (task accuracy) as a function of iterations in INLP. The high PN feature (word-overlap) is more difficult to remove (noisier probing accuracy), and is accompanied by drop in the task accuracy.}
    \label{fig:inlp_all}
\end{figure}

\begin{figure}[t]
    \centering
    \includegraphics[scale=0.54]{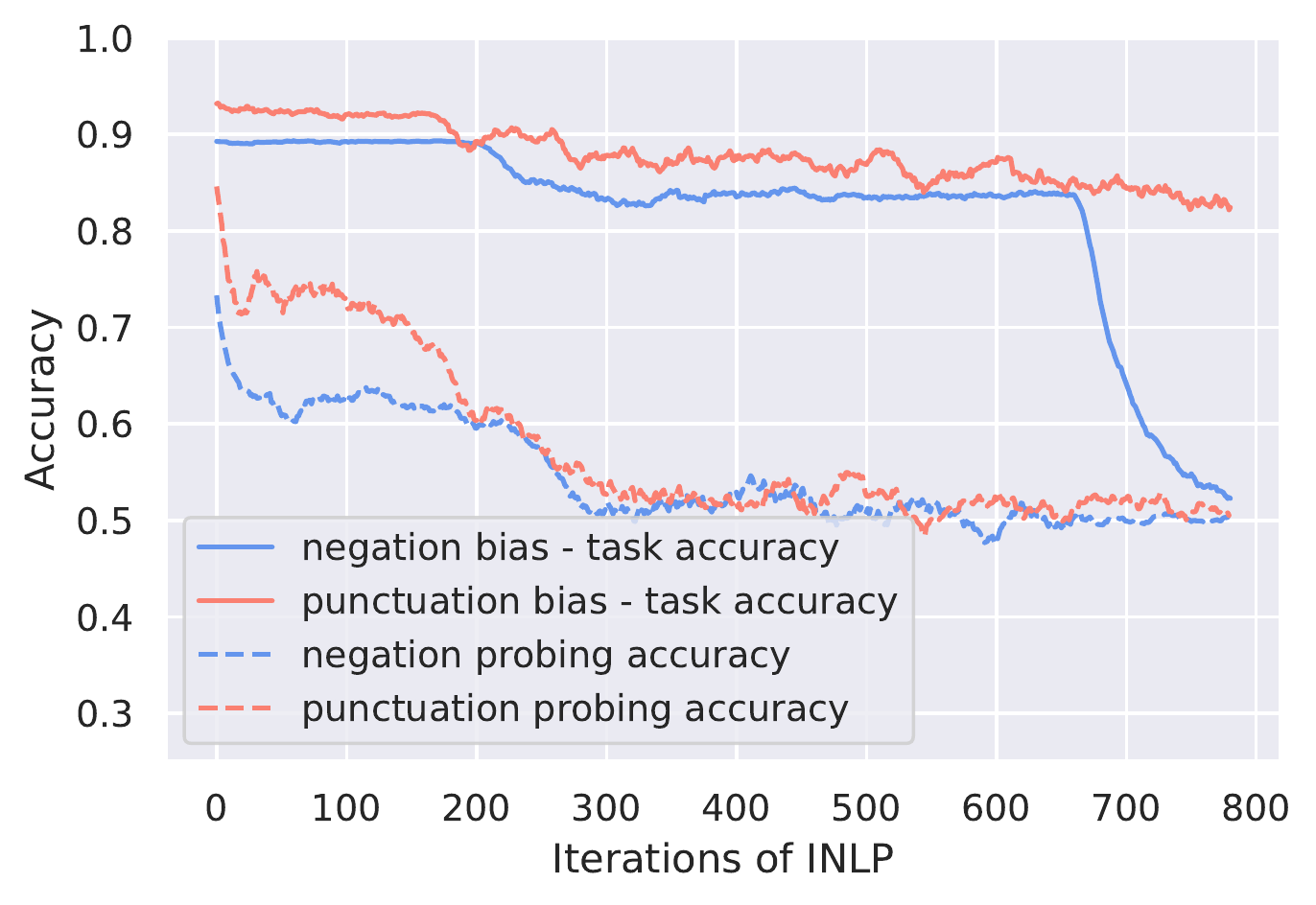}
    \caption{Extractability of the spurious feature (probing accuracy) and the main task accuracy (task accuracy) as a function of iterations in INLP. The high PN feature (negation) is more difficult to remove (noisier probing accuracy), and is accompanied by drop in the task accuracy.}
    \label{fig:inlp_neg}
\end{figure}

\begin{table*}[t]
\begin{small}
    \centering
    \begin{tabular}{ccccccccc}
        \toprule
                   & \multicolumn{2}{c}{Negation-bias} & \multicolumn{2}{c}{Word-overlap bias} & \multicolumn{2}{c}{Synthetic-NLI}                               \\
                   & $\mathcal{C}$                     & Acc.                                  & $\mathcal{C}$                     & Acc. & $\mathcal{C}$ & Acc. \\
        \midrule
        Baseline   & 2.6                               & 86.7                                  & 3.5                              & 90.5 & 47.6          & 100  \\
        Subsampling \cite{sagawa2020investigation} & 2.6                               & 87.8                                  & 3.6                               & 91.5 & \cellcolor{red!74}{10.2}          & 97.7 \\
        POE \cite{karimi-mahabadi-etal-2020-end}       & \cellcolor{cyan!20}{2.8}                               & 88.9                                  & \cellcolor{cyan!60}{4.2}                               & 91.3 & \cellcolor{red!9.4}{42.9}       &  99.9     \\
        DFL \cite{karimi-mahabadi-etal-2020-end}       & \cellcolor{cyan!30}{2.9}                               & 89.2                                  & \cellcolor{cyan!30}{3.9}                               & 88.3 & \cellcolor{cyan!1.8}{48.5}              & 100     \\
        Group-DRO ~\cite{Sagawa*2020Distributionally}  & \cellcolor{cyan!20}{2.8}                               & 89.8                                  & \cellcolor{cyan!90}{4.7}                               & 91.5 & \cellcolor{red!66}{14.7}          & 100  \\
        \bottomrule
    \end{tabular}
    \caption{Extractability of the spurious feature for various robust training methods. In general, the representation is more invariant to the feature if it has low PN (synthetic NLI) than if it has high PN (negation and word-overlap bias).
    }
    \label{tab:probing_extended}
\end{small}
\end{table*}

\paragraph{Training Details} For INLP, we use the 1024 dimensional representation of the first token from RoBERTa-Large as the representation of the input. The linear model is trained and evaluated on subsets of the dataset where the probing label is balanced.

In Figure~\ref{fig:inlp} we observed that for the lexical overlap spurious correlation, the performance for the main task drops significantly on the minority examples. Here, we show that we also observe a decrease in the average performance albeit less than that for the minority group. One potential explanation for why we observe larger drop on the minority examples is that learning an invariant representation leads the model to solve the easier examples in the majority group (e.g. high lexical overlap examples with entailment label) at the cost of the minority examples. The performance for the main task on all dev examples for lexical overlap is shown in Figure ~\ref{fig:inlp_all}. We additionally also compare to the negation spurious correlation which also has a type 2 dependency in Figure \ref{fig:inlp_neg} --- we observe that the main task accuracy remains much higher than that for lexical overlap but eventually drops down suddenly.

\section{Encoding of Spurious Feature: Extended Results}
\label{sec:encoding_extended}

In addition to the results reported for lexical overlap and synthetic bias in NLI, we also verify the hypothesis for negation spurious correlation and evaluate Group-DRO \cite{Sagawa*2020Distributionally} on all spurious correlations in Table~\ref{tab:probing_extended}.



\end{document}